\documentclass[aps,prc,twocolumn,floatfix,showpacs,a4paper,nofootinbib,amsmath,amssymb]{revtex4-1}
\usepackage{graphicx}
\usepackage{color}
\usepackage{dcolumn}
\usepackage{bm}
\usepackage{relsize}
\newcommand{\be}{\begin{equation}}
\newcommand{\ee}{\end{equation}}
\newcommand{\ba}{\begin{eqnarray}}
\newcommand{\ea}{\end{eqnarray}}
\newcommand{\bd}{\begin{displaymath}}
\newcommand{\ed}{\end{displaymath}}

\begin{document}

\title{Quantitative Entropy Study of Language Complexity}
\author{R.R. Xie$^1$, W.B. Deng$^1$, D.J. Wang$^2$ and L.P. Csernai$^3$}
\affiliation{
$^1$  Complexity Science Center, Institute of Particle Physics, Central China Normal University, Wuhan 430079, China\\
$^2$ School of Science, Wuhan University of Technology, Wuhan 430070, China\\
$^3$ Institute of Physics and Technology, University of Bergen, Allegaten 55, 5007 Bergen, Norway }

\date{\today}
\begin{abstract}
We study the entropy of Chinese and English texts, based on characters 
in case of Chinese texts and based on words for both languages. Significant
differences are found between the languages and between different
personal styles of debating partners. The entropy analysis points in the 
direction of lower entropy, that is of higher complexity. Such a text
analysis would be applied for individuals of different styles, a 
single individual
at different age, as well as different groups of the population.
\end{abstract}


\maketitle

\section{Introduction}

Sustainable development was first addressed by Erwin Schr\"odinger
\cite{Schroedinger}
based on entropy, where development was characterized by increasing
"orderliness" (nowadays complexity). He pointed out that the development
of highly complex forms of matter (or life) should be built on
less complex forms. This means decreasing entropy, while increasing
the entropy of matter should be avoided if we want to maintain 
sustainable development.

Recently the build up of complexity on the example of 1 kg matter in 
different forms was studied quantitatively, starting from the 
simplest example of ideal gases, 
and then continue with more complex chemical, biological, and living 
structures \cite{CSV2016}. 
The complexity of these systems was assessed quantitatively, 
based on their entropy. We use the method introduced in 
Ref.  \cite{CSV2016}, which attributed the same 
entropy to known physical systems and to complex organic molecules
up to most complex Human Genome DNA. 

Schr\"odinger \cite{Schroedinger} has also discussed and concluded that
the emergence of life does not require new fundamental laws of physics,
which allow for non-increasing entropy. Actually, as the Earth is
an open system \cite{CsPSX2016}, 
with a boundary condition strongly decreasing
entropy, this boundary condition enforces development towards decreasing
entropy, i.e. increasing complexity.

The Human brain has a vastly superior possibility of complexity, than 
biological molecules, and it carries abstract information, as well as 
many vegetative and reflex functions. The direct calculation of
the complexity of the coding in the Human neural network is beyond
our present knowledge, but we can make studies of the stored,
consciously reachable, information and its complexity.

The conscious thinking can be indirectly studied via the analysis of
Human languages. We can think about one subject at a time, just like
we can speak about one subject at a time.

\section{Language Complexity}

As discussed in Ref.  \cite{CSV2016}, to analyze the system from 
entropy or complexity point of view
we have to consider two basic aspects: (i) the quantum of information
or of the substance we analyse and (ii) the possibility of all
configurations in a set of degrees of freedom, as well as the
realized, realizable or existing configurations from the set of
all possible configurations. 

Regarding the first point (i) in
physics we quantized the phase space (the six dimensional position and
momentum space) and have introduced the volume of the phase space
element based on the quantum mechanical uncertainty relation.

In case of a language the basic element could be the word. This can 
also be the basic element of the conscious thinking. At this time
we do not have sufficient information on how a ``word" is represented
in a neural network, how many neurons and synapses are involved,
and what is the weight of the corresponding material. Hopefully
in the future we can acquire the knowledge to reply to these questions.
This situation is similar to the early development of statistical
physics, when kinetic theory and thermodynamics were already known, 
with entropy and the second law of thermodynamics. At this time it was
already realized that the phase space should be quantized, but
before the quantum mechanics one did not know what should be
this phase space volume. This did lead to a state where entropy was 
only defined up to a constant, which could be chosen free. 
Still similar systems could be analysed quantitatively, and compared to 
each other.

When we chose the word as the quantum of a language we are in a
similar situation as the early thermodynamics. 
A constant is {\it remaining to be determined}, 
to compare the entropy of the
language to that of the ideal gases or the Human DNA sequence.

The second condition (ii) is not very problematic in case of a language,
the given amount of words can be determined by analysis of texts.
Then the number of all possible configurations can be calculated
for any given length sentence. For long sentences this number of
configurations can become astronomically high, but one can analyse
the distribution of the lengths of sentences as well as the maximal
length. This will make the number of possible configurations 
finite. Subsequently one can analyse existing texts and can
evaluate the number of existing configurations. This last step,
can be done for a single person's language (who wrote extensively,
so that we can analyse his or her language). It can be done for
writings in a region where the language is used, or for all users of 
the language.\footnote{
In some languages the computational analysis of texts may be 
problematic, e.g., in Hungarian the form of a word in a given
text is changing to the extent that it is not possible to find
the root of a word in a dictionary. The good knowledge of the
language and grammar would be necessary to do this analysis, what 
computational analysis programs cannot do at this time.}

\section{Analysis of Chinese Language with Characters}

As a first example we use the analysis of the Chinese language to have
an order of magnitude estimate of the quantitative complexity or
entropy of Human thinking via the language.

The Chinese language uses characters. On average a person uses about
3000 characters in communication. The characters may form words of
one, two or more characters. These afterwards, form sentences, which are 
separated by periods (and exclamation or question marks) in writing.

Texts of about 26000-80000 Chinese characters were analysed in 
four samples, Sample $I$ to $IV$
\cite{c1,c2,c3,c4}.
We evaluated how many different 
Chinese characters were contained in a given sample, $N_c$. 
Then in the first evaluation, we checked how many 
one  character sentences, $N^c_1$,
two  character sentences, $N^c_2$,
three character sentences, $N^c_3$, and so on,
up to 35 character sentences were in the samples. See table \ref{t1}.

\begin{table}[h]   
\setlength{\tabcolsep}{2.1pt}
\renewcommand{\arraystretch}{1.25}
\begin{tabular}{crrrrrrrrrrr} \hline\hline \phantom{\Large $^|_|$}\!\!
Sample\!\!\!\!\! & $N_s$&$N_c$ & $N^c_1$ & $N^c_2$ & $N^c_3$ & $N^c_4$ &
	$N^c_5$ & $N^c_6$ & $N^c_7$ & $N^c_8$ & $N^c_9$ \\
\hline
$I$   & 79959& 2553&163& 375& 248& 225& 209& 193& 168& 195& 149\\
$II$  & 79470& 2137& 69& 130& 100& 126& 123& 181& 170& 156& 169\\
$III$ & 26671& 2096&  4&   4&   5&   6&  24&  20&  32&  27&  30\\
$IV$  & 29083& 1916&  1&   4&   5&  20&  19&  18&  29&  38&  48\\
\hline \hline
\end{tabular}
\caption{
Number of all Chinese characters, $N_s$, and of different 
Chinese characters, $N_c$, in the Sample texts, $I$ to
$IV$ are shown.
Then the sentences (separated by periods) are analysed:
the one character sentences, two character sentences,
and so on.
The number of different $k$-character sentences, $N^c_k$ were
counted in the sample texts.  The longest sentences were between
162, 119, 145, and 129 characters for the four samples respectively.}
\label{t1}
\end{table}   

Let us now consider the two character sentences. This can be formed
by choosing one character of the $N_c$ for the first position,
and another from the $N_c$ for the second position. The two characters 
may be identical and the sequence of the characters is meaningful.
Consequently the maximum number of possible 
{\it two character sentences} 
is $N_c^2$, and the probability of one configuration is $p_i = 1/N_c^2$.
Thus the maximum entropy of all possible two character sentences
is
\ba
H(X_2^{max}) &=& - \sum_{all}  p_i \ln p_i = 
- N_c^2  \frac{1}{N_c^2}  \ln \frac{1}{N_c^2}
\nonumber \\
&=& \ln N_c^2 =  15.690, 15.334, 15.296, 11.116\ ,
\nonumber \\
\label{HX2max}
\ea
for Samples $I,\, II,\, III,\, IV,$ respectively.

In real physical or biological situations not all 
(hypothetical) configurations are 
realized. The
number of observed or Realized (R) different 
{\it two character sentences} 
for Sample $I$  is only $N^c_2= 375$. Consequently the corresponding
 specific configuration entropy is
\ba
H(X_2^R) &=& - \sum_{i=1}^{N^c_2}  p_i \ln p_i = 
- N^c_2  \frac{1}{N_c^2}  \ln \frac{1}{N_c^2} 
\nonumber \\
&=& 
2  N^c_2 \ln(N_c) /  N_c^2 = 9.027\cdot 10^{-4} .
\label{HX2}
\ea
This entropy is proportional with the number of two-character sentences, 
$N^c_2$. At the same time  $N^c_2$ is also proportinal with the 
size of the Sample text.

We can do the same analysis in this Sample $I$ text,
for sentences of one, three, four, etc., ..., 162 Chinese
character sentences. These are unrelated configurations 
and as the entropy is additive the specific entropy of
{\it all sentences} of Sample text $I$, based on 
Chinese characters is
\ba
\sigma_c &=& H(X_1^R) + H(X_2^R) + H(X_3^R) + H(X_4^R) + ... 
\nonumber \\
&=& 5.009\cdot 10^{-1} +9.027\cdot 10^{-4} + 3.508 \cdot 10^{-7} + ...
\nonumber \\
&=& 5.018\cdot10^{-1}.
\nonumber \\
\ea
One can see that the few (one, two, three) character sentences
provide the largest contribution to the entropy, and the 
longer ones have minor contribution. The higher level of complexity
is achieved by minimizing the use of one or two character sentences.
The very long sentences have very large number of hypothetical
possibilities, while occur very seldom in the text. 
The contribution of 10 character sentences to the entropy is
$\sigma_{10c} < 10^{-30}$, and the longer ones are even smaller.
One could take into account the relative frequencies of the 
different length sentences, but the relative frequencies of long 
sentences in the sample texts is also rapidly decreasing.
Therefore their entropy contribution is utterly negligible. 
This also indicates
the hint that the length of the Sample text is not very important
beyond some number as it leads to relatively small change
in the results.

\begin{table}[h]   
\setlength{\tabcolsep}{2.5pt}
\renewcommand{\arraystretch}{1.25}
\begin{tabular}{cccc} \hline\hline \phantom{\Large $^|_|$}\!\!
Sample\ \  & $N_s$& $\sigma_{c}$ & $\sigma_{c/10k}$ \\
\hline
$I$   &  79959\ \ & $5.018 \cdot10^{-1}$\ \  & $6.276 \cdot 10^{-2}$ \\
$II$  &  79470\ \ & $2.480 \cdot10^{-1}$\ \  & $3.121 \cdot 10^{-2}$\\
$III$ &  26671\ \ & $1.461 \cdot10^{-2}$\ \  & $5.477 \cdot 10^{-3}$\\
$IV$  & 29083\ \ & $3.961 \cdot10^{-3}$\ \  & $1.362 \cdot 10^{-3}$\\
\hline \hline
\end{tabular}
\caption{
Specific entropy of the Sample texts based on Chinese characters, where $N_s$ 
is the number of characters in the Sample text, $\sigma_{c}$ is the entropy
of the text, and $\sigma_{c/10k}$ is the entropy
of the text normalized to 10000 character length.}
\label{sc1}
\end{table}   

The entropy of a Sample text is proportional to the length of the text. 
In order to
compare texts of different lengths we can introduce a specific entropy
for 10000 characters (or words), so for Sample $I$:
\be
\sigma_{c/10k} \equiv  10000 \cdot \sigma_c / N_s = 
 = 6.276 \cdot 10^{-2}\ .
\ee

Samples $I-IV$ have a different texts 
with different parameters. The entropy 
analysis can be preformed the same way as for Sample $I$, resulting:
\be
\sigma_{c/10k} = 
(6.276,\        3.121,\      0.5477,\    0.1362)  \cdot 10^{-2} \ , 
\ee
for Sample texts $I-IV$ respectively. The shorter text samples 
have a tendency to 
give smaller length normalized specific entropy. The results are 
summarized in Table
\ref{sc1}.

\section{Analysis of Chinese Language with Words}

In the Chinese language, although single characters may 
correspond to a word, certain two or three character combinations
are unique and can be considered as words. So in this sense
words can be considered as the basic parts of a sentence
instead of Chinese characters. See table \ref{t2}.

\begin{table}[h]   
\setlength{\tabcolsep}{2.0pt}
\renewcommand{\arraystretch}{1.25}
\begin{tabular}{crrrrrrrrrrr} \hline\hline \phantom{\Large $^|_|$}\!\!
Sample\!\!\!\!\! & $N_s$&$N_w$ & $N^w_1$ & $N^w_2$ & $N^w_3$ & $N^w_4$ &
	$N^w_5$ & $N^w_6$ & $N^w_7$ & $N^w_8$ & $N^w_9$ \\
\hline
$I$   & 49835&10122&558& 304& 348& 279& 282& 283& 249& 257& 241\\
$II$  & 47911& 8169&208& 174& 219& 268& 260& 264& 273& 262& 261\\
$III$ & 16780& 5086&  5&  13&  22&  38&  46&  41&  53&  55&  54\\
$IV$  & 18501& 4775&  4&  13&  19&  55&  50&  47&  68&  65&  49\\
\hline \hline
\end{tabular}
\caption{
Number of all Chinese words, $N_s$, and the different
Chinese words, $N_w$, in the Sample texts, $I-IV$ are shown.
Then the sentences (separated by periods) are analysed:
the one word sentences, two word sentences,
and so on.
The number of different $k$-word sentences, $N^w_k$ were
counted in the sample texts.\\
}
\label{t2}
\end{table}   

In Chinese
writing the words are not separated by spaces, but commas, quotation 
marks and other punctuation may separate words. We employ the 
package "jiebaR" with {\it R language} to distinguish the words.  

We can calculate the maximum specific entropy for
all hypothetical $k$-word combinations, using the number of different
Chinese words in the sample text. For example for
{\it two word sentences}
\ba
H(X_2^{max}) &=& - \sum_{all}  p_i \ln p_i = 
- N_w^2 \frac{1}{N_w^2}  \ln \frac{1}{N_w^2}
\nonumber \\
&=& \ln N_w^2 =  18.444, 18.016,  17.068, 14.192\ ,
\nonumber \\
\label{HX2max}
\ea
for Samples $I,\, II,\, III,\, IV,$ respectively. These are
smaller than the max entropies for Chinese characters as the 
observed number of words is smaller than the number of characters.

The number of observed 
{\it two word sentences} in the Sample texts is of course much 
smaller, than the hypothetical maximum, thus the specific entropy for 
two word sentences is also smaller:
\be
H(X_2^R) =
(508.4 ,\ 
 229.4 ,\ 
  8.390,\ 
  7.096) \cdot 10^{-3}.
\ee
for Samples $I,\, II,\, III,\, IV,$ respectively.

Then we add up the entropy contribution of 
{\it all observed sentences}
of all lengths in the Sample texts. 
This provides the total specific entropy 
\be
\sigma_w = 
(508.5,\ 
  229.4,\
   8.399,\
   7.106 )\
 \cdot 10^{-3},\
\ee
for Sample texts $I-IV$ respectively. We  summarize these data 
in Table \ref{sw1}.

\begin{table}[h]   
\setlength{\tabcolsep}{2.5pt}
\renewcommand{\arraystretch}{1.25}
\begin{tabular}{cccc} \hline\hline \phantom{\Large $^|_|$}\!\!
Sample\ \  & $N_s$& $\sigma_{w}$ & $\sigma_{w/10k}$ \\
\hline
$I$   &  49835\ \ & $5.085 \cdot10^{-1}$\ \  & $1.020 \cdot 10^{-1}$ \\
$II$  &  47911\ \ & $2.294 \cdot10^{-1}$\ \  & $4.788 \cdot 10^{-2}$\\
$III$ &  16780\ \ & $8.399 \cdot10^{-3}$\ \  & $5.005 \cdot 10^{-3}$\\
$IV$  & 18501\ \ & $7.106 \cdot10^{-3}$\ \  & $3.841 \cdot 10^{-3}$\\
\hline \hline
\end{tabular}
\caption{
Specific entropy of the Sample texts based on Chinese words, where $N_s$ 
is the number of words in the Sample text, $\sigma_{w}$ is the entropy
of the text, and $\sigma_{w/10k}$ is the entropy
of the text normalized to 10000 word length.}
\label{sw1}
\end{table}   

The Entropies obtained from the analysis of the words is similar to those 
that were based on characters. The difference between the entropy and the 
length normalized entropy based on characters and words is smaller 
in the case of words.

\section{Analysis of English texts}

The analysed English text samples 
\cite{e1,e2,e3,e4},
contained 102613, 93668, 6480, and 8992 words. The 3rd and 4th texts
are from the first presidential candidacy debate of Hillary Clinton and 
Donald Trump. See table \ref{t3}.

\begin{table}[h]   
\setlength{\tabcolsep}{2.1pt}
\renewcommand{\arraystretch}{1.25}
\begin{tabular}{crrrrrrrrrr} \hline\hline \phantom{\Large $^|_|$}\!\!
Sample\!\!\!\!\! &$N_s$&$N_w$ & $N^w_1$ & $N^w_2$ & $N^w_3$ & $N^w_4$&
	$N^w_5$ & $N^w_6$ & $N^w_7$ & $N^w_8$  \\
\hline
$I$  &102613& 7966&  0&  6& 33&  68& 155& 120&  189& 272\\
$II$ & 93668& 5745& 44& 10& 12&  11&   7&   8&   28&  35\\
$III$&  6480& 1309&  2&  6&  7&  19&  29&  28&   22&  28\\
$IV$ &  8992& 1225& 18& 12& 27&  42&  74&  76&   66&  64\\
\hline \hline
\end{tabular}
\caption{
Number of all English words, $N_s$, and of 
different English words, $N_w$, in the Sample texts.
Then the sentences (separated by periods) are analysed:
the two word sentences, three word sentences
and so on.
The number of different $k$-word sentences, $N^w_k$ were
counted in the sample texts.
While Samples $I$ and $II$ are extended written texts, $III$ and $IV$ are
debates of Hillary Clinton and Donald Trump, respectively. The debate
texts are shorter and thus also their vocabulary is more constrained.
}
\label{t3}
\end{table}   

Noticeable that while most Chinese sentences have 10 words or less
the in the analysed English text most sentences have about 20 words!
This has an interesting effect on the complexity or entropy analysis
of the text.

\begin{figure}[h]     
\begin{center}
\resizebox{0.95\columnwidth}{!}
{\includegraphics{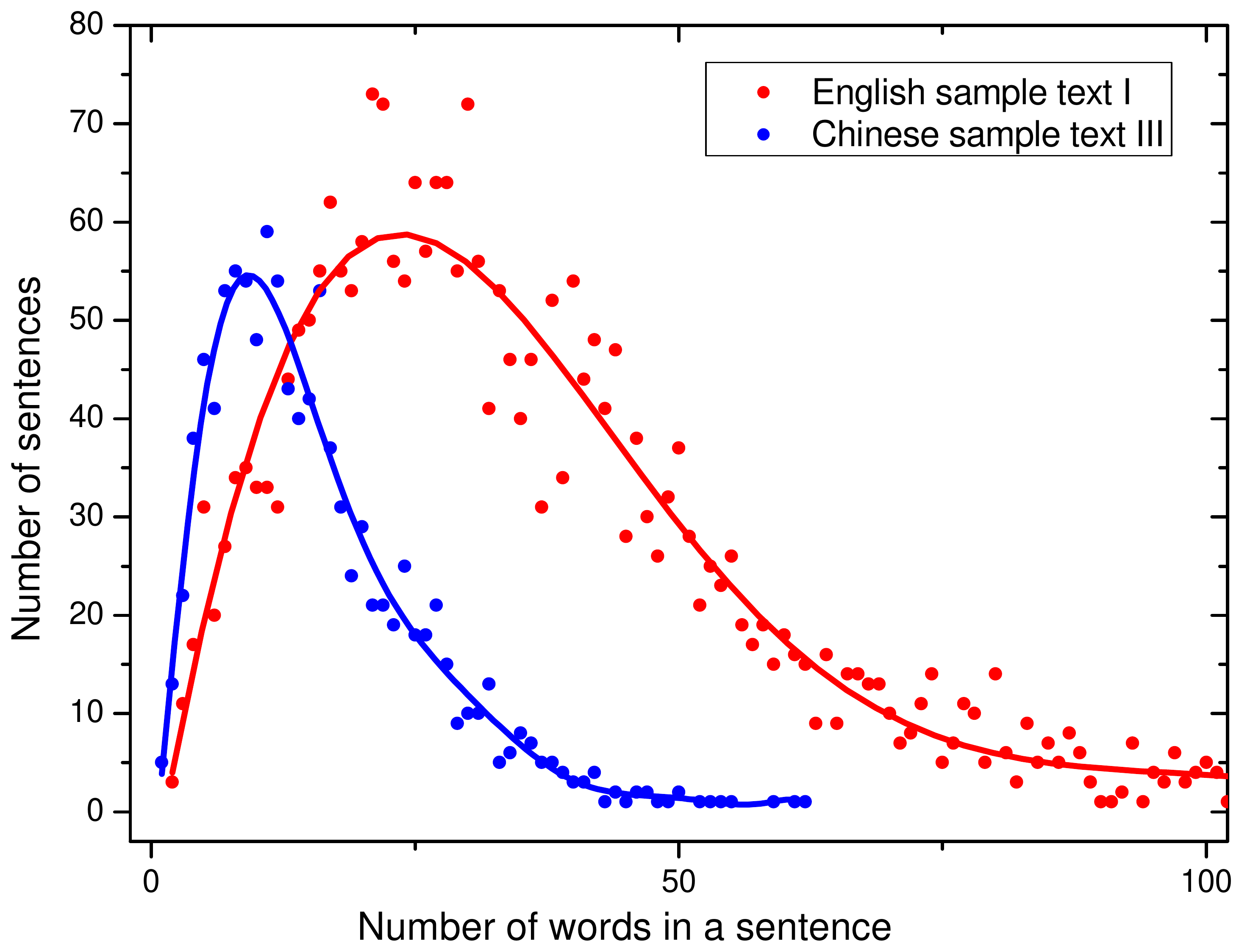}}
\vskip 1cm
\caption{(Color online)
The distribution of the sentences according to their length.
The length is measured by the number of words in a sentence, while
the number of sentences of a given length in a Sample text is shown.
The red dots correspond to the English Sample text I, peaking at 
$\sim$ 26 words, while the
blue dots to the Chinese Sample text III, peaking at $\sim$ 8 words.
The lines are to guide the eye.
}
\label{f2}
\end{center}
\end{figure}

We can calculate the maximum specific entropy for
all hypothetical $k$-word combinations. For example for
{\it two word sentences}
\be
H(X_2^{max})= 
 17.876,\  17.312,\  14.354,\  14.222,
\ee
for the three English Sample texts. This is
larger than the max entropies for Chinese word texts, due to the
larger number of words in the text.  See Figure \ref{f2}.

The number of observed two word sentences in the text is of course much 
smaller, than the hypothetical maximum, thus the specific entropy for 
realized {\it two word sentences} is also smaller
\be
H(X_2^R) =
8.494  \! \cdot 10^{-7}\!,\
5.245  \! \cdot 10^{-6}\!,\
5.026  \! \cdot 10^{-5}\!,\
1.137  \! \cdot 10^{-4}\!,\
\ee
for the English text Sample.

Then we add up the entropy contribution of all observed sentences
of all lengths. This provides the total specific entropy for 
{\it all sentences}
for the four English Sample texts. See Table \ref{se1}.

\begin{table}[h]   
\setlength{\tabcolsep}{2.5pt}
\renewcommand{\arraystretch}{1.25}
\begin{tabular}{cccc} \hline\hline \phantom{\Large $^|_|$}\!\!
Sample\ \  & $N_s$& $\sigma_{w}$ & $\sigma_{w/10k}$ \\
\hline
$I$  & 102613\ \ & $8.499 \cdot10^{-7}$\ \  & $8.283 \cdot 10^{-8}$ \\
$II$  &  93668\ \ & $6.630 \cdot10^{-2}$\ \  & $7.078 \cdot 10^{-3}$\\
$III$ &    6480\ \ & $1.102 \cdot10^{-2}$\ \  & $1.700 \cdot 10^{-2}$\\
$IV$  &   8992\ \ & $1.046 \cdot10^{-1}$\ \  & $1.163 \cdot 10^{-1}$\\
\hline \hline
\end{tabular}
\caption{
Specific entropy of the Sample texts based on English words, where $N_s$ 
is the number of words in the Sample text, $\sigma_{w}$ is the entropy
of the text, and $\sigma_{w/10k}$ is the entropy
of the text normalized to 10000 word length.}
\label{se1}
\end{table}   

\begin{figure}[h]     
\begin{center}
\resizebox{0.95\columnwidth}{!}
{\includegraphics{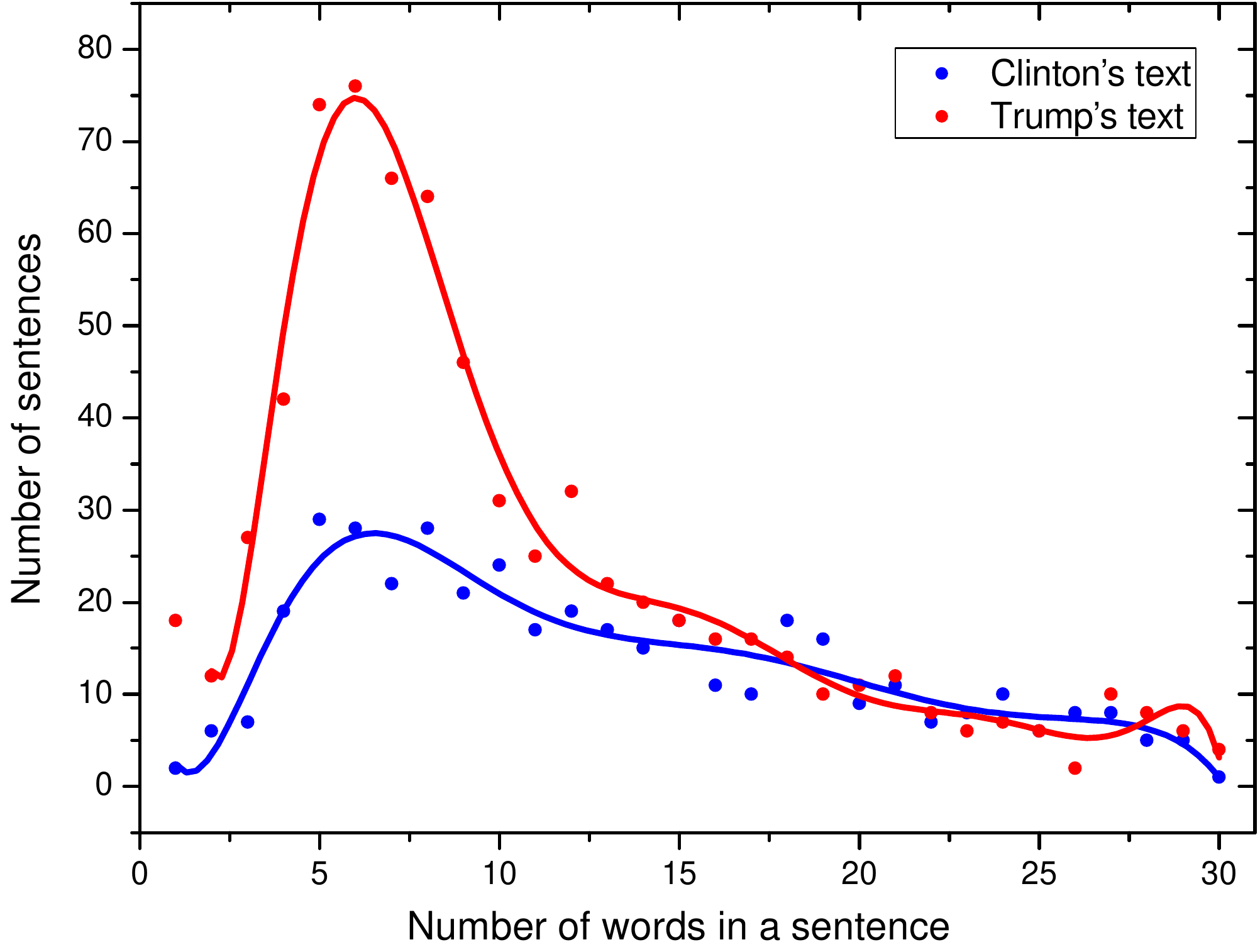}}
\vskip 1cm
\caption{(Color online)
The distribution of the sentences according to their length for
the first presidential debate between Hillary Clinton and
Donald Trump.
The red dots correspond to Trump's text, while the
blue dots to Clinton's. Trump's text is dominated by short
sentences peaking with 76 sentences of 6 word length.
The lines are to guide the eye.
}
\label{f3}
\end{center}
\end{figure}

For the 1st Sample text, 
this is the same as that of the
shortest $2$-word sentences in the text, because the next longer
$3$-word sentences have an entropy value that is 4 orders of magnitude 
smaller. Due to the lack of single word sentences, and the 
small number of two word sentences the entropy of the English text is
much smaller than that of the Chinese texts. 
In the much shorter debate texts of Clinton and Trump the number of 
very short sentences dominates. Trump has a much larger number of 
short sentences and this increases the total entropy of his
text, contrary to the fact that the number of the words in his
text is significantly larger. See Figure \ref{f3}.

\section{Discussion}

The analysis of the complexity or entropy of languages can of course be
used for comparing different languages, or different texts, or authors
to each other. There is a vast amount of literature analysing languages
with many different methods. Here we have chosen a relatively simple,
and transparent method.

But the entropy value as a general feature of material can actually 
lead to conclusion regarding the entropy of the physical and biological
structure of the brain, and the information content in abstract sense.
The language can be representative of the conscious operation of the 
brain. The physical and biological complexity has to be much larger, 
as the brain is responsible for the vegetative operation of the 
nervous system as well as the dynamical changes of the operation and
the human activity.
The language itself is just a static set of information, but it has
to be learned, so it is a structure, which is the product of training
or learning. The language itself can characterize the development,
see e.g. \cite{JYun2015}.

The language can also be attributed to a given amount of material.
It is a given part of the brain, even if we cannot identify it.
Plausibly the same part of the brain carries other static information
as well as dynamical information also. This way in addition to the
specific entropy of the language, $\sigma_c$ or $\sigma_w$ we can
also estimate the physical entropy $S_{1kg}$ or at least a lower
limit of it.

In case of usual (Shannon) entropy estimates the normalization is not the 
same as the physical one, but it is perfectly sufficient for comparative
studies of these type of structural entropies. 

In this analysis the role of physical phase-space or configuration space
is taken over by the ``word-space" or ``Chinese character-space". These
spaces could in principle be extended to infinity, but in fact taking
all words of a language in a historical period the word-space of a 
language is finite. This is also necessary as the language is a 
means of communication. Thus, we cannot add up the word-space
of all languages.

\section{Conclusions}

We have demonstrated quantitatively the increasing complexity of materials,
and used the entropy for unit amount of material to be able to get
a measure. This idea stems from Ervin Schroedinger, but our
knowledge today makes it possible to extend the level of 
quantitative discussion to complex live materials.

We may continue these studies to higher levels of material structures,
like living species, artificial constructions, symbiotic coexistence of
different species, or grouping of the same species. up to even structures
in Human society.

The main achievement of the earlier  work 
\cite{CSV2016}
was to show how the entropy in the physical
phase space and the entropy of structural degrees of freedom(Shannon entropy) 
can be discussed on the same platform.  
For further developments it is important to point out two
fundamental aspects of the entropy concept: 
(i) the {\it quantization} of the 
space of a given degree of freedom, and 
(ii) the {\it selection of the realized, realizable
or beneficial configurations} from all the possible ones. 

In the present work we introduced a quantization as the number of words
or Chinese characters. At this moment of time we do not
know how to relate this quantization to the basic physical quantum
of the occupation of an elementary phase-space element. Thus,
the relative normalization of the quantitative complexity or entropy
of the language is still missing. We would need a much more detailed
knowledge about the representation of language in the neural network
of the human brain.

The other aspect of the entropy calculation is actually solved
in case of the human language or languages, as the realized configurations
can be relatively easily determined by the analyses of the
texts.

The Sample text examples presented are all static point.
As we see on the example of the nervous system the dynamical 
change of the entropy of the system is also important. The 
text analysis could trace down the change of the complexity
of the texts of an individual, which could be a measure of the
period of increasing complexity at early years compared to 
decreasing complexity and increasing entropy later.
Such analysis could be performed on the novels of authors, who were
active for many years.
The dynamics and
direction of these changes is also essential as 
shown in Ref. \cite{PCs1980}.

\section*{Acknowledgements}

This research was partially supported by the 
Academia Europaea Knowledge Hub, Bergen, by the  
Institute of Advanced Studies, K\H{o}szeg, 
by the National Natural Science Foundation of China (Grant No. 11505071),
and by the Wuhan University of Technology.



\begin{thebibliography}{99}

		
\bibitem{Schroedinger}	
Erwin Schr\"odinger: {\it What is life? - The Physical 
Aspect of the Living Cell}, (The Cambridge University Press, 1944)
Based on the Lectures delivered under the auspices of the Trinity
College, Dublin, in February 1943

\bibitem{CSV2016}
L.P. Csernai, S.F. Spinnangr, S. Velle,
Quantitative assessment of increasing complexity,
arXiv: 1609.04637

\bibitem{CsPSX2016}
L.P. Csernai, I. Papp, S.F. Spinnangr and Yilong Xie,
Physical Basis of Sustainable Development,
Journal of Central European Green Innovation, {\bf 4}, 39-50 (2016).

\bibitem{c1}
Lau Shaw: {\it	Lao Zhang's Philosophy}	(1928)

\bibitem{c2}
Lau Shaw: {\it	Nameless Highland has a Name} (1955)

\bibitem{c3}
Eileen Chang: 
{\it Crumbs of Ligumaloes - the First Incense Burnt} (1943)

\bibitem{c4}
	Eileen Chang: {\it Chuang Shi Ji (Fata Morgana)} (1945)

\bibitem{e1}
Jonathan Swift: {\it Gulliver's Travels} (1726)

\bibitem{e2}
Charles Robert Darwin: {\it The Origin of Species} (1859)

\bibitem{e3}
Hillary Clinton: {\it 1st Presidential debate} (26/09/2016)

\bibitem{e4}
Donald Trump: {\it 1st Presidential debate} (26/09/2016)

\bibitem{JYun2015}
J. Yun, P.-J. Kim, H. Jeong,
Anatomy of Scientific Evolution,
PLoS ONE {\bf 10}(2): e0117388  (2015).
doi:10.1371/journal.pone.0117388.

\bibitem{PCs1980}
L. P\'enzes, and L.P. Csernai, 
\"Uber den Zusammenhang von Lebensdauer, Konstitution und Information;
Zeitschrift f\"ur Alternsforschung, {\bf 35}, 285-296 (1980).








\end{thebibliography}
\end{document}